\pdfoutput=1

\documentclass[11pt]{article}

\usepackage[final]{acl}

\usepackage{times}
\usepackage{latexsym}

\usepackage{amssymb}
\usepackage{pifont}
\usepackage{hyperref}
\usepackage{xurl}

\usepackage{tabularx} 

\usepackage[T1]{fontenc}

\usepackage[utf8]{inputenc}

\usepackage{microtype}
\usepackage{lipsum}
\usepackage{inconsolata}

\title{A Dual-Prompting for Interpretable Mental Health Language Models}

\author{Hyolim Jeon$^{1,2}$\thanks{~~Equal contribution.},~Dongje Yoo$^{3*}$,~Daeun Lee$^{1}$,\\~\textbf{Sejung Son$^{1}$},~\textbf{Seungbae Kim$^{4}$},~\textbf{Jinyoung Han$^{1,2}$}\thanks{~~Corresponding author.} \\
  $^1$Department of Applied Artificial Intelligence, Sungkyunkwan University, Seoul, South Korea \\
  $^2$Department of Human-AI Interaction, Sungkyunkwan University, Seoul, South Korea \\
  $^3$Department of Computer Engineering, Chung-Ang University, Seoul, South Korea \\
  $^4$Computer Science \& Engineering Department, University of South Florida, Tampa, FL, USA \\
  \texttt{\{gyfla1512,~maze0717\}@g.skku.edu}
  \texttt{\{delee12,~jinyounghan\}@skku.edu} \\
  \texttt{pass120cau.ac.kr, seungbae@usf.edu} \\}

\usepackage{inconsolata}

\usepackage{booktabs}
\usepackage{xcolor}
\usepackage{pifont}
\usepackage{multirow}
\usepackage{enumitem}
\usepackage{array}
\usepackage{amsmath}
\usepackage{subcaption}
\usepackage{graphicx}
\usepackage{makecell}
\usepackage{cleveref}

\begin{document}

\maketitle

\begin{abstract}
Despite the increasing demand for AI-based mental health monitoring tools, their practical utility for clinicians is limited by the lack of interpretability.
The CLPsych 2024 Shared Task\footnote{\url{https://clpsych.org/shared-task-2024/}} aims to enhance the interpretability of Large Language Models (LLMs), particularly in mental health analysis, by providing evidence of suicidality through linguistic content. 
We propose a dual-prompting approach: (i) Knowledge-aware evidence extraction by leveraging the expert identity and a suicide dictionary with a mental health-specific LLM; and (ii) Evidence summarization by employing an LLM-based consistency evaluator. 
Comprehensive experiments demonstrate the effectiveness of combining domain-specific information, revealing performance improvements and the approach's potential to aid clinicians in assessing mental state progression.

\end{abstract}


\section{Introduction}
The global healthcare system faces significant challenges from mental health conditions such as depression and suicidal ideation~\cite{darrudi2022challenges}, emphasizing the need for an advanced monitoring system for early intervention~\cite{galea2020mental}.

In response, NLP researchers have paid attention to identifying mental states, often leveraging social media data~\cite{chen-etal-2023-detection, liu2023improving, lee2023towards}. Notably, the most recent development involves the application of Large Language Models (LLMs), which have demonstrated robust capabilities in general language processing in mental health analysis~\cite{yang2023towards, yang2023mentalllama, xu2023leveraging}. 
Specifically, \citet{amin2023will} conducted a comparison of ChatGPT's zero-shot capability in identifying suicide and depression, contrasting it with previous methods that relied on previous Pre-trained Language Models (PLMs). Furthermore, \citet{lamichhane2023evaluation} evaluated ChatGPT's effectiveness in recognizing stress, depression, and suicide, emphasizing its strong grasp of language in texts related to mental health.

However, these studies have focused on identifying mental health status through a black box model, posing a challenge in interpreting the rationale behind their outcomes~\cite{schoene2023example, zhang2022natural}. Accordingly, efforts have been made to enhance the interpretability of mental health analysis, such as guiding LLMs to emphasize emotional cues~\cite{yang2023towards} and developing open-source LLMs by training them with data from mental health-related social media~\cite{yang2023mentalllama}. Nevertheless, a lack of reliability still remains; recent LLMs are often unreliable or inconsistent~\cite{agrawal2022large}, potentially due to the lack of mental health-related knowledge~\cite{yang2023towards}. This problem has significantly delayed the practical use of LLMs in clinical settings~\cite{malhotra2024xai}. 

To address this issue, the CLPsych 2024 Shared Task~\cite{chim2024overview} introduces the challenge of utilizing open-source LLMs to enhance their interpretability in mental health analysis, specifically focusing on detecting suicidality through linguistic content in social media data. Particularly, the shared task includes two subtasks: (i) \textit{Task A} requires finding key phrases from each post to support suicide risk, and (ii) \textit{Task B} aims to provide a summary of evidence related to the user's suicide risk across multiple posts. 

In this paper, we design an enhanced prompt for the extraction task (Task A) by assigning an expert identity, enabling LLMs to function as an expected agent~\cite{xu2023expertprompting}, and leveraging a suicide dictionary~\cite{lee2022detecting} to capture suicide-related context. Here, we utilize mental health-specific LLM, MentaLLaMA~\cite{yang2023towards}.
For the summarization task (Task B), we employ a consistency evaluator~\cite{luo2023chatgpt} to improve the consistency of outcomes with multiple summaries.

The extensive experiments illustrate that combining domain-specific information with few-shot learning enhances the extraction of evidence, resulting in an improvement in recall from 91.0\% to 92.2\%. 
Additionally, our findings indicate that an LLM trained with general datasets is more effective in mitigating hallucination in summarization tasks than a domain-specific LLM.
We believe our approach can support clinicians in assessing mental state progression.

\begin{figure}[t!]
    \centering
    \includegraphics[width=1\linewidth]{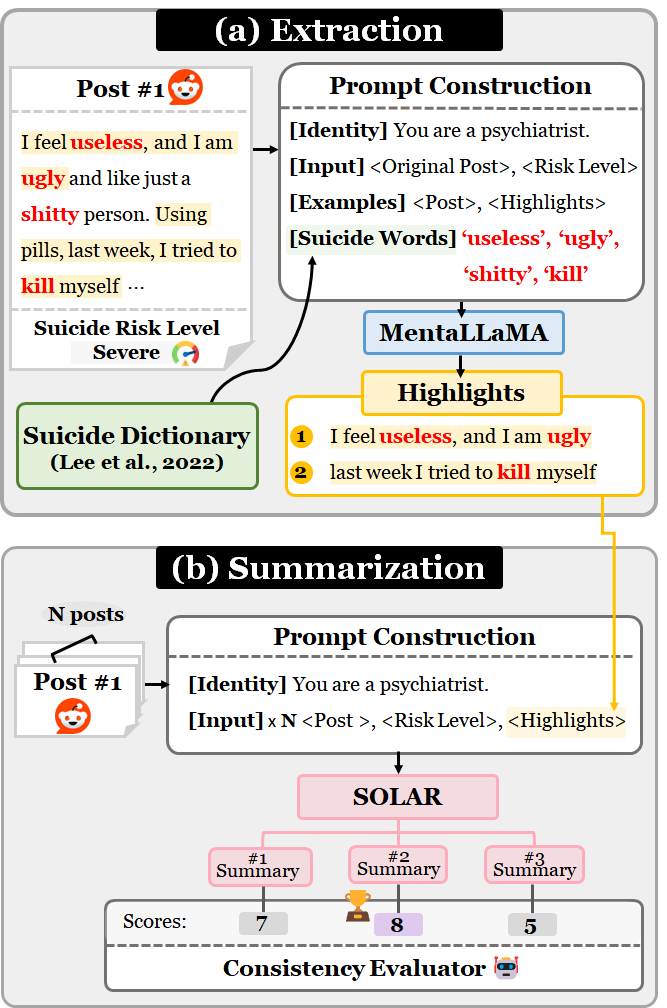}
    \vspace{-0.2in}
    \caption{The overall architecture of the proposed approach: (a) Knowledge-aware Evidence Extraction (\S \ref{sec:task1}) and (b) Evidence Summarization with LLM-based Consistency Evaluator (\S \ref{sec:task2})}
    \label{fig:model}
    \vspace{-0.2in}
\end{figure}

\section{Methodology}
Our aim is (i) to extract evidence supporting the user's suicide risk from each post and (ii) to summarize all the evidence across multiple posts. To this end, we design two prompting strategies to instruct LLMs for trustworthy reasoning in mental health analysis. These strategies include Knowledge-aware Evidence Extraction (\S \ref{sec:task1}) and Evidence Summarization with an LLM-based Consistency Evaluator (\S \ref{sec:task2}). The overall proposed approach is depicted in Figure~\ref{fig:model}, and the full text of each prompt is available in Appendix~\ref{sec:appendix}.


\subsection{Knowledge-aware Evidence Extraction}~\label{sec:task1}
As shown in Figure~\ref{fig:model}(a), the prompt for the extraction task includes the original post and the assigned user's suicide risk level. For few-shot learning, we incorporate examples that are not included in the evaluation dataset.
Moreover, we apply the three prompting strategies to address the unreliability issue of LLMs arising from the lack of mental health-related knowledge~\cite{yang2023towards}.


\noindent\textbf{A. Mental health-specific LLM.} 
In order to tackle the zero-shot challenge in LLMs~\cite{han2023medalpaca}, we utilize MentaLLaMA-chat-13B~\cite{yang2023mentalllama}, fine-tuned with 105K mental health-related social media data, demonstrating its efficacy in mental health-related tasks.

\noindent\textbf{B. Assigning expert identity.} 
As LLMs tend to provide insight into their cognitive processes when assigning predefined roles~\cite{li2023camel,xu2023expertprompting}, we employ prompts to allocate the domain expert identity (e.g., \textit{`You are a psychiatrist'}). 


\begin{table}[]
\centering
\caption{Example of suicide words~\cite{lee2022detecting}.}
\label{tab:dic_dataset}
\resizebox{\linewidth}{!}{\begin{tabular}{lcl}
\hline
\textbf{Suicide Risk}    & \textbf{\# of Words} & \textbf{Examples} \\ \hline \hline
Low      & 48                   & emptiness, overthink \\ \hline
Moderate & 83                   & psychiatric, pain     \\ \hline
Severe & 111 & cutting, die \\ \hline
\end{tabular}}
\end{table}

\noindent\textbf{C. Utilizing a suicide dictionary.} 
Since a domain-specific dictionary can aid LLMs in capturing relevant context~\cite{yang2023towards}, we utilize a suicide dictionary~\cite{lee2022detecting}, which has proven effective in identifying suicidal ideation on social media data.
As shown in Table~\ref{tab:dic_dataset}, the dictionary uses the UMD Reddit Suicidality Dataset~\cite{shing2018expert}, comprising 279 words validated by domain experts. 
If the given post includes words from the suicide dictionary~\cite{lee2022detecting}, the model identifies and incorporates these words into our prompt, instructing the LLM to consider these words attentively.

\subsection{Evidence Summarization with LLM-based Consistency Evaluator}~\label{sec:task2}
As shown in Figure~\ref{fig:model}(b), the prompt for the summarization incorporates multiple posts and the assigned user's suicide risk level. Additionally, an expert identity is assigned, similar to the previous step.
However, despite the advancements in LLMs, hallucination and inconsistency still remain significant concerns~\cite{tang-etal-2023-understanding, tang2023evaluating}. To mitigate this issue, we apply the two following strategies that can enhance consistency.

\noindent\textbf{A. Extract-then-Generate.} 
\citet{zhang-etal-2023-extractive-summarization} demonstrated the effectiveness of prompts that incorporate an extractive summary for abstractive summarization. Following this approach, the proposed prompt integrates the extracted phrases obtained from the preceding step (\S\ref{sec:task1}). The full text of each prompt is available in Appendix~\ref{sec:summary-prompt}.

\noindent\textbf{B. Consistency Evaluator.} 
We adopt a consistency evaluator proposed by~\citet{luo2023chatgpt}. Initially, multiple candidate answers are generated through the LLM. We then compute consistency scores (ranging from 1 to 10) for each candidate, assessing the extent to which the generated summary aligns with the original posts, utilizing the consistency evaluator.
In the end, the answer with the highest score from multiple candidates is selected as the final result.
Here, we adopt SOLAR~\cite{kim2023solar} as the summarizer and evaluator, known for its recent outstanding performance\footnote{\url{https://huggingface.co/spaces/HuggingFaceH4/open_llm_leaderboard}}. Further details comparing summarizer and evaluator are provided in \S\ref{sec:ex-then-gen}. 
\begin{table}[]
\centering
\caption{Statistics of the evaluation dataset.}
\label{tab:data_statistics}
\resizebox{\linewidth}{!}{%
\begin{tabular}{c|cc|cc}  
\textbf{} & \multicolumn{2}{c|}{\textbf{Highlights}} & \multicolumn{2}{c}{\textbf{Summarization}} \\ \hline
\textbf{Suicide Risk} & \multicolumn{2}{c|}{\textbf{\# posts (avg. \# length)}} & \multicolumn{2}{c}{\textbf{\# users (avg. \# posts)}} \\  \hline \hline
\textbf{Low} & \multicolumn{2}{c|}{{17 (1,149)}} & \multicolumn{2}{c}{{13 (1.31)}} \\ 
\textbf{Moderate} & \multicolumn{2}{c|}{{91 (1,132)}} & \multicolumn{2}{c}{{75 (1.21)}} \\
\textbf{Severe} & \multicolumn{2}{c|}{{54 (1,178)}} & \multicolumn{2}{c}{{37 (1.46)}} \\ \hline
\textbf{Total} & \multicolumn{2}{c|}{\textbf{162 posts}} & \multicolumn{2}{c}{\textbf{125 users}} \\ \hline 
\end{tabular}}
\end{table}

\section{Experiments}
\subsection{Evaluation Dataset}~\label{sec:dataset}
The CLPsych 2024 shared task~\cite{chim2024overview} provides the UMD Reddit Suicidality Dataset~\cite{zirikly2019clpsych,shing2018expert}, consisting of 79,569 posts from 37,083 subreddits by 866 Reddit users who posted on r/SuicideWatch between 2008 and 2015. Each user in the dataset is assigned a label that indicates the severity of suicidality (i.e., No, Low, Moderate, or Severe), determined by crowdsourcers and domain experts.

The evaluation dataset comprises a subset of users labeled with \textit{Low}, \textit{Moderate}, and \textit{Severe} risks validated by domain experts. It includes 162 posts distributed among 125 users and
the statistics of the dataset are summarized in Table~\ref{tab:data_statistics}. 


\subsection{Experimental Settings}
All experiments are conducted on a GeForce RTX 3090 Ti GPU with 26GB of memory. To minimize the memory cost of 16-bit weights, we employ the bitsandbytes library~\cite{dettmers2022gptint}, converting them to int8 using vector-wise quantization~\cite{dettmers2022optimizers} without significant quality loss. Each prompt is processed independently to mitigate the impact of dialogue history.

\subsection{Evaluation Metrics} 
Note that the ground truth dataset was not provided to the participants. Therefore, all the evaluation metrics and reported results are supplied by the organizers of the CLPsych 2024 Shared Task. 

\noindent\textbf{(1) Similarity:} BERTScore~\cite{zhang2019bertscore} is employed for the extraction task to assess token similarity using contextual embeddings. 

\noindent\textbf{(2) Consistency:} For the summarization task, a natural language inference (NLI) model~\cite{laurer2024less} is applied to assess the consistency of individual sentences in the provided evidence summary. Specifically, the contradiction scores are calculated between the predicted outcomes and each ground truth summary sentence. The resulting sentence-level consistency score is then determined as 1 minus the probability of the contradiction prediction.

\begin{table}[]
\centering
\caption{Comparison of performance between zero-shot and few-shot learning for the extraction task (Task A) using the evaluation dataset.}
\label{tab:results}
\resizebox{0.8\linewidth}{!}{%
\begin{tabular}{cccc}
\multicolumn{1}{c}{\multirow{2}{*}{\textbf{Model}}} & \multicolumn{3}{c}{\textbf{Highlights (Task A)}} \\ \cline{2-4}
\multicolumn{1}{c}{} & Pre.↑ & Rec.↑ & F1↑ \\ \hline \hline
\multicolumn{1}{c|}{Ours w/ Zero-shot} & \textbf{0.913} & {0.910} & 0.911 \\
\multicolumn{1}{c|}{Ours w/ Few-shot} & {0.912} & \textbf{0.922} & \textbf{0.917} \\ \hline
\end{tabular}%
}
\end{table}

\section{Results \& Analysis} \label{sec:analysis}
To demonstrate the effectiveness of the proposed method, we compare its performance with various approaches and conduct the case study where our proposed approach performs better. Note that due to the absence of ground truth from the organizer, quantitative analysis was limited, leading us to focus on qualitative analysis instead.
Additionally, we manually paraphrase any examples from the data to preserve user anonymity.

\subsection{Analysis on Knowledge-aware Evidence Extraction}~\label{sec:analysis_task_a}
Table~\ref{tab:results} shows the results of our approach on the evaluation dataset, with precision, recall, and F1 scores, for the highlights task (Task A).

\noindent\textbf{Analysis on few-shot learning. } 
We find an improvement in recall from 91.0\% to 92.2\% by integrating few-shot learning. This suggests the importance of providing examples for few-shot learning in domain-specific tasks, particularly in clinical settings~\cite{han2023medalpaca}.

\noindent\textbf{Analysis on suicide-dictionary. } 
We find integrating a suicide dictionary~\cite{lee2022detecting} also improves domain knowledge in extracting evidence. Specifically, it allows thorough consideration of suicide risk factors that might be overlooked due to their general meaning, such as \textit{`family'} and \textit{`credit'}, which have been validated by domain experts as suicide-related words. Examples of the results are provided below.
 
\begin{quote}
    \textbf{\textit{Response w/ Suicide Dictionary:}} [``\textcolor{red}{Fear} of failing.","\textcolor{red}{Fear} of hurting.''], 
    [``\textcolor{red}{working} as of \textcolor{red}{credit} problems.''],
    [``Don't want to be a burden or face my \textcolor{red}{friends} and \textcolor{red}{family}.'']
\end{quote}

\noindent\textbf{Analysis on expert identity.}~\label{sec:analysis_task_diverse_expert}
We explore the performance of the LLM by employing different expert identities, such as psychology, counseling, and psychiatry. This analysis aims to understand how the model's behavior varies depending on the assigned role. For example, when the role is assigned as a psychologist, the LLM tends to prioritize the user's negative self-perception (e.g., \textit{`ugly'} and \textit{`hate'}) to a greater extent. Conversely, adopting the identity of a counselor enables the model to focus on the relationship (e.g.,\textit{ `broke up'} and \textit{`divorce'}), which may contribute to feelings of isolation. Additionally, we observe that assigning a psychiatrist role is likely to focus on clinical markers, such as emotional distress (e.g., \textit{`anxiety'}) and history of abuse (e.g., \textit{`assaulted'}), which can be connected to suicidal ideation. Hence, we suggest that selecting an appropriate identity aligned with the research objective can offer valuable insights.

\begin{quote}
    \textbf{\textit{Response w/ Psychology Identity:} } 
    [``I am \textcolor{red}{ugly}, I am \textcolor{red}{annoying}, I am \textcolor{red}{unwanted}''],
    [``I \textcolor{red}{hate} me'']
    
    \textbf{\textit{Response w/ Counselor Identity :}} 
    [``Fuck, we \textcolor{red}{broke up} three weeks ago''],
    [``\textcolor{red}{disconnected} from everybody'']
    
    \textbf{\textit{Response w/ Psychiatry Identity:}} 
    [``I will never go to school because of my \textcolor{red}{depression}.''],
    [``I am feeling \textcolor{red}{anxious/angry} and constantly \textcolor{red}{lonely}''],
    [``When I was 4 years old, I was sexually \textcolor{red}{abused}'']
\end{quote}



\subsection{Analysis on Evidence Summarization with LLM-based Consistency Evaluator}
\noindent\textbf{Analysis on Extract-then-Generate. }\label{sec:ex-then-gen}
We explore the efficacy of incorporating extractive summaries from Task A for the evidence summarization task. We observe that the hallucination issue frequently arises when extractive summaries are absent. This indicates that our approach enhances consistency by providing contextual information~\cite{zhang-etal-2023-extractive-summarization}. 
For a better understanding, we provide an example below. We notice that the LLM misinterprets the expression \textit{`wishing to do it'} as a desire for success, resulting in generating \textit{`self-distrust in achievements'} by the LLM.

\begin{quote}
    \textbf{\textit{Posts:} } I was \textcolor{red}{thinking about} when I tried to hang myself, \textcolor{red}{wishing to do it} now. \\
    \textbf{\textit{Response w/o Extract-then-Generate:} } They exhibit risk due to \textcolor{red}{cognitions} (\textcolor{red}{self-distrust} in \textcolor{red}{achievements}).
\end{quote}

\begin{figure}
    \centering
    \includegraphics[width=0.65\linewidth]{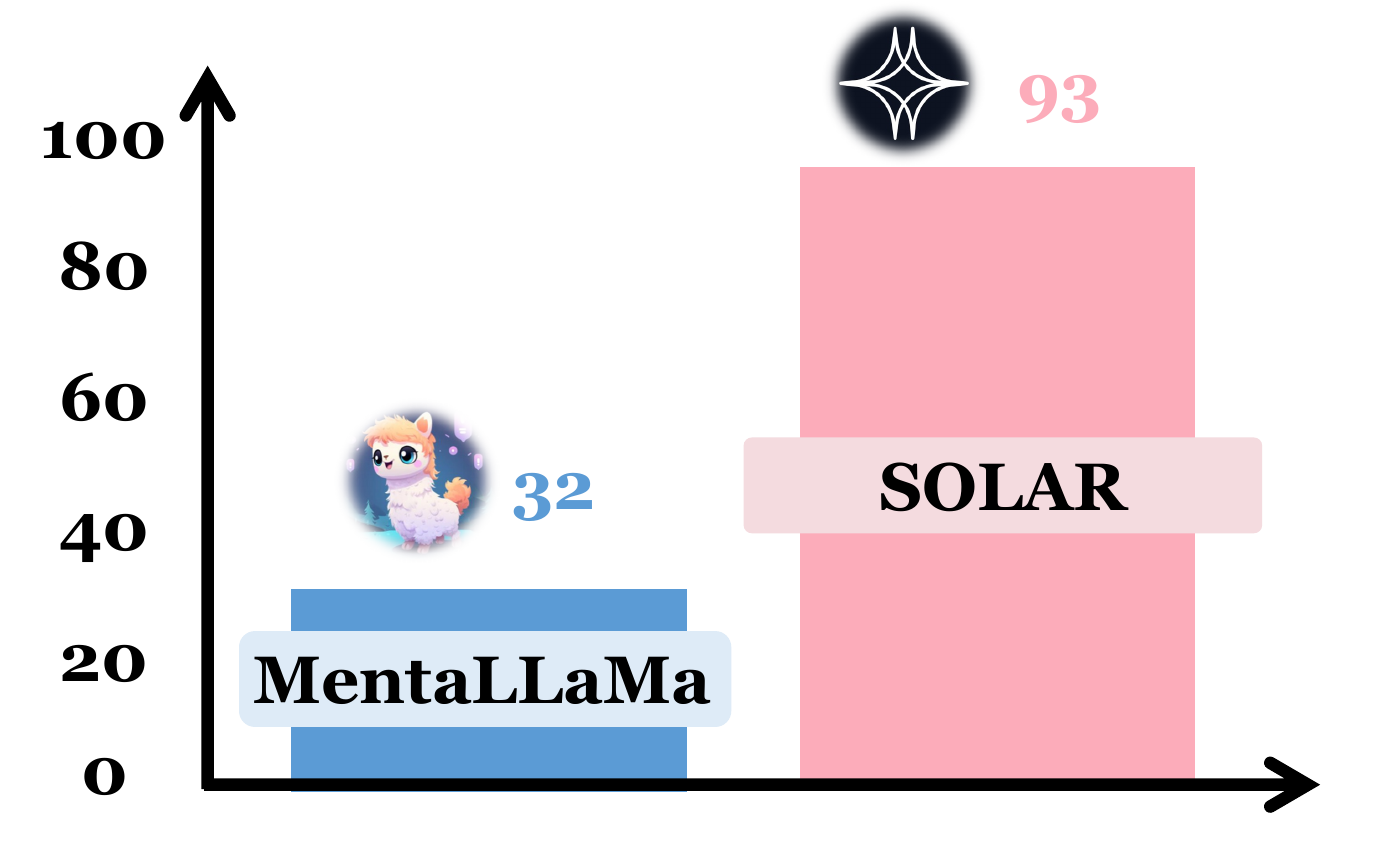}
    \caption{Winner count comparison for MentaLLaMA~\cite{yang2023mentalllama} and SOLAR~\cite{kim2023solar} in 125 evaluation dataset using evaluator.}
    \label{fig:mhvssolar}
\end{figure}

\noindent\textbf{Comparison LLMs with Consistency Evaluator. }
Table~\ref{tab:result_consistency} shows the results of our approach on the evaluation dataset, along with the mean consistency scores for the summarization task (Task B).
We find that using only SOLAR~\cite{kim2023solar} as a summarizer performed better than using both SOLAR~\cite{kim2023solar} and MentaLLaMA~\cite{yang2023mentalllama}. This also can be found in Figure~\ref{fig:mhvssolar}, when we use both summarizers, the evaluator selects 93 results from SOLAR~\cite{kim2023solar} and 32 from MentaLLaMA~\cite{yang2023mentalllama} as the final outputs from the 125 evaluation set. 
This implies that domain-specific models tend to perform worse than general LLMs, like ChatGPT~\cite{luo2023chatgpt} or SOLAR~\cite{kim2023solar}, on general linguistic tasks such as abstractive summarization~\cite{wu2023bloomberggpt}. 
Moreover, MentaLLaMA~\cite{yang2023mentalllama} exhibits biased hallucination issues by generating mental-health-related words like `\textit{stuck}' or `\textit{bother}' regardless of original contexts, leading to inconsistency.
In future work, we plan to explore the comparison of evaluators and summarizers using a broader range of LLMs to gain additional insights.


\begin{table}[]
\centering
\caption{Comparison of performance among different summarizers for the summarization task (Task B) using the evaluation dataset.}
\label{tab:result_consistency} 
\resizebox{\linewidth}{!}{%
\begin{tabular}{cccc}
\multicolumn{1}{c}{\multirow{2}{*}{\textbf{Summarizer}}} & \multicolumn{3}{c}{\textbf{Summarization (Task B)}} \\ \cline{2-4}
\multicolumn{1}{c}{} & \multicolumn{3}{c}{Consistency↑} \\ \hline \hline
\multicolumn{1}{c|}{SOLAR \& MentaLLaMa} & \multicolumn{3}{c}{0.970} \\
\multicolumn{1}{c|}{SOLAR} & \multicolumn{3}{c}{\textbf{0.973}} \\ \hline
\end{tabular}%
}
\end{table}

\begin{quote}
    \textbf{\textit{Posts:} } If I couldn't return, I would \textcolor{red}{jump on the train}, or my dad \textcolor{red}{wouldn't take me} to the TV show ... \\
    \textbf{\textit{Response w/ MentaLLaMa:}} The user shows a feeling of being \textcolor{red}{stuck} and \textcolor{red}{bothered} by others.
\end{quote} 


\subsection{Error Analysis}
While our proposed approach demonstrates outstanding performance, there are a few cases where the model fails to recognize crucial evidence supporting the suicide risk level and extracts sentences that are irrelevant to the potential suicide risk. Concerning practical utility, the lack of reliability has considerably impeded the implementation of LLMs in clinical settings~\cite{malhotra2024xai}.

\begin{quote}
    \textbf{\textit{Response w/ Expert Identity:} } [``This subreddit is a \textcolor{red}{fantastic} place.'']

    \textbf{\textit{Response w/ Suicide Dictionary:} } [``I \textcolor{red}{love} everyone in this subreddit.'']  
\end{quote}

\section{Conclusion}
In this study, we introduced promising prompting strategies that can provide evidence supporting suicide risk levels on social media data. We enhanced the LLM interpretability by incorporating domain-specific elements like assigning a psychiatrist identity and combining a suicide word. Additionally, we improved the consistency in summarization by using a consistency evaluator with multiple candidates. The proposed dual-prompting approach provides reliable reasoning, making it suitable for monitoring mental health-related risks.

\section*{Limitations}
Since ground truth is not provided, quantitative comparisons are limited. Therefore, we rely on qualitative comparisons, which may be subjective.
Our experiments use only the smallest version of LLMs due to limited resources.
Providing inferences about suicidality using social media data is inherently subjective, allowing for various interpretations among researchers~\cite{keilp2012suicidal}. Moreover, the experimental data may be sensitive to demographic and media-specific biases~\cite{hovy2016social}. While the effectiveness of leveraging social media data for mental health analysis may be constrained in specific clinical settings~\cite{ernala2019methodological}, adopting a practical model promises the potential to discern diverse statistical patterns and biases across various objectives~\cite{jacobson2020ethical}. Although the suicide dictionary~\cite{lee2022detecting} has demonstrated effectiveness in predicting suicide risk, its reliance on social media data for construction might restrict its generalizability. 
Furthermore, the dictionary was constructed using the same dataset as the one utilized in the Shared Task, which is anticipated to introduce a certain degree of bias.

\noindent\textbf{Future Work.} In future work, we plan to explore a wider range of prompt templates to enhance overall performance further. For instance, the prompts could be diversified by applying various LLM-based consistency evaluators, including ChatGPT and LLaMa2.
Our ultimate objective is to expand the scope to cover diverse mental health domains, such as depression and bipolar disorder, and validate its effectiveness comprehensively. To achieve this, we plan to investigate domain-specific fine-tuning methods for LLMs in the mental health field, thereby extending the model to a more interpretable context.

\section*{Ethical Statement} 
The CLPsych 2024 shared task~\cite{chim2024overview} prioritized responsible data utilization by providing exclusive access to the dataset for researchers aligned with ethical principles. Consequently, all task participants must adhere to data use agreements and ethical practices during the competition.
Our research strongly emphasizes ethics, particularly in (i) protecting the privacy of Reddit users and (ii) preventing potential misuse of the dataset. We strictly adhered to Reddit's privacy policy\footnote{\url{https://www.reddit.com/policies/privacy-policy}} to ensure user anonymity~\cite{benton2017ethical,williams2017towards}.

\section*{Acknowledgments}
We commit to acknowledging the assistance of the American Association of Suicidology in making the dataset available, in any publications.

This research was supported by the Ministry of Education of the Republic of Korea and the National Research Foundation (NRF) of Korea (NRF-2022S1A5A8054322) and the International Research \& Development Program of the National Research Foundation of Korea (NRF) funded by the Ministry of Science and ICT (RS-2023-00265683).

\bibliography{reference}

\begin{thebibliography}{37}
\expandafter\ifx\csname natexlab\endcsname\relax\def\natexlab#1{#1}\fi

\bibitem[{Agrawal et~al.(2022)Agrawal, Hegselmann, Lang, Kim, and Sontag}]{agrawal2022large}
Monica Agrawal, Stefan Hegselmann, Hunter Lang, Yoon Kim, and David Sontag. 2022.
\newblock Large language models are few-shot clinical information extractors.
\newblock In \emph{Proceedings of the 2022 Conference on Empirical Methods in Natural Language Processing}, pages 1998--2022.

\bibitem[{Amin et~al.(2023)Amin, Cambria, and Schuller}]{amin2023will}
Mostafa~M Amin, Erik Cambria, and Bj{\"o}rn~W Schuller. 2023.
\newblock Will affective computing emerge from foundation models and general ai? a first evaluation on chatgpt.
\newblock \emph{IEEE Intelligent Systems}, 38:2.

\bibitem[{Benton et~al.(2017)Benton, Coppersmith, and Dredze}]{benton2017ethical}
Adrian Benton, Glen Coppersmith, and Mark Dredze. 2017.
\newblock Ethical research protocols for social media health research.
\newblock In \emph{Proceedings of the first ACL workshop on ethics in natural language processing}, pages 94--102.

\bibitem[{Chen et~al.(2023)Chen, Zhang, Wu, and Zhu}]{chen-etal-2023-detection}
Siyuan Chen, Zhiling Zhang, Mengyue Wu, and Kenny Zhu. 2023.
\newblock \href {https://doi.org/10.18653/v1/2023.emnlp-main.562} {Detection of multiple mental disorders from social media with two-stream psychiatric experts}.
\newblock In \emph{Proceedings of the 2023 Conference on Empirical Methods in Natural Language Processing}, pages 9071--9084, Singapore. Association for Computational Linguistics.

\bibitem[{Chim et~al.(2024)Chim, Tsakalidis, Gkoumas, Atzil-Slonim, Ophir, Zirikly, Resnik, and Liakata}]{chim2024overview}
Jenny Chim, Adam Tsakalidis, Dimitris Gkoumas, Dana Atzil-Slonim, Yaakov Ophir, Ayah Zirikly, Philip Resnik, and Maria Liakata. 2024.
\newblock Overview of the clpsych 2024 shared task: Leveraging large language models to identify evidence of suicidality risk in online posts.
\newblock In \emph{Proceedings of the Ninth Workshop on Computational Linguistics and Clinical Psychology}. Association for Computational Linguistics.

\bibitem[{Darrudi et~al.(2022)Darrudi, Khoonsari, and Tajvar}]{darrudi2022challenges}
Alireza Darrudi, Mohammad Hossein~Ketabchi Khoonsari, and Maryam Tajvar. 2022.
\newblock Challenges to achieving universal health coverage throughout the world: a systematic review.
\newblock \emph{Journal of preventive medicine and public health}, 55(2):125.

\bibitem[{Dettmers et~al.(2022{\natexlab{a}})Dettmers, Lewis, Belkada, and Zettlemoyer}]{dettmers2022gptint}
Tim Dettmers, Mike Lewis, Younes Belkada, and Luke Zettlemoyer. 2022{\natexlab{a}}.
\newblock \href {https://openreview.net/forum?id=dXiGWqBoxaD} {{GPT}3.int8(): 8-bit matrix multiplication for transformers at scale}.
\newblock In \emph{Advances in Neural Information Processing Systems}.

\bibitem[{Dettmers et~al.(2022{\natexlab{b}})Dettmers, Lewis, Shleifer, and Zettlemoyer}]{dettmers2022optimizers}
Tim Dettmers, Mike Lewis, Sam Shleifer, and Luke Zettlemoyer. 2022{\natexlab{b}}.
\newblock 8-bit optimizers via block-wise quantization.
\newblock \emph{9th International Conference on Learning Representations, ICLR}.

\bibitem[{Ernala et~al.(2019)Ernala, Birnbaum, Candan, Rizvi, Sterling, Kane, and De~Choudhury}]{ernala2019methodological}
Sindhu~Kiranmai Ernala, Michael~L Birnbaum, Kristin~A Candan, Asra~F Rizvi, William~A Sterling, John~M Kane, and Munmun De~Choudhury. 2019.
\newblock Methodological gaps in predicting mental health states from social media: triangulating diagnostic signals.
\newblock In \emph{Proceedings of the 2019 chi conference on human factors in computing systems}, pages 1--16.

\bibitem[{Galea et~al.(2020)Galea, Merchant, and Lurie}]{galea2020mental}
Sandro Galea, Raina~M Merchant, and Nicole Lurie. 2020.
\newblock The mental health consequences of covid-19 and physical distancing: the need for prevention and early intervention.
\newblock \emph{JAMA internal medicine}, 180(6):817--818.

\bibitem[{Han et~al.(2023)Han, Adams, Papaioannou, Grundmann, Oberhauser, L{\"o}ser, Truhn, and Bressem}]{han2023medalpaca}
Tianyu Han, Lisa~C Adams, Jens-Michalis Papaioannou, Paul Grundmann, Tom Oberhauser, Alexander L{\"o}ser, Daniel Truhn, and Keno~K Bressem. 2023.
\newblock Medalpaca--an open-source collection of medical conversational ai models and training data.
\newblock \emph{arXiv preprint arXiv:2304.08247}.

\bibitem[{Hovy and Spruit(2016)}]{hovy2016social}
Dirk Hovy and Shannon~L Spruit. 2016.
\newblock The social impact of natural language processing.
\newblock In \emph{Proceedings of the 54th Annual Meeting of the Association for Computational Linguistics (Volume 2: Short Papers)}, pages 591--598.

\bibitem[{Jacobson et~al.(2020)Jacobson, Bentley, Walton, Wang, Fortgang, Millner, Coombs~III, Rodman, and Coppersmith}]{jacobson2020ethical}
Nicholas~C Jacobson, Kate~H Bentley, Ashley Walton, Shirley~B Wang, Rebecca~G Fortgang, Alexander~J Millner, Garth Coombs~III, Alexandra~M Rodman, and Daniel~DL Coppersmith. 2020.
\newblock Ethical dilemmas posed by mobile health and machine learning in psychiatry research.
\newblock \emph{Bulletin of the World Health Organization}, 98(4):270.

\bibitem[{Keilp et~al.(2012)Keilp, Grunebaum, Gorlyn, LeBlanc, Burke, Galfalvy, Oquendo, and Mann}]{keilp2012suicidal}
John~G Keilp, Michael~F Grunebaum, Marianne Gorlyn, Simone LeBlanc, Ainsley~K Burke, Hanga Galfalvy, Maria~A Oquendo, and J~John Mann. 2012.
\newblock Suicidal ideation and the subjective aspects of depression.
\newblock \emph{Journal of affective disorders}, 140(1):75--81.

\bibitem[{Kim et~al.(2023)Kim, Park, Kim, Lee, Song, Kim, Kim, Kim, Lee, Kim et~al.}]{kim2023solar}
Dahyun Kim, Chanjun Park, Sanghoon Kim, Wonsung Lee, Wonho Song, Yunsu Kim, Hyeonwoo Kim, Yungi Kim, Hyeonju Lee, Jihoo Kim, et~al. 2023.
\newblock Solar 10.7 b: Scaling large language models with simple yet effective depth up-scaling.
\newblock \emph{arXiv preprint arXiv:2312.15166}.

\bibitem[{Lamichhane(2023)}]{lamichhane2023evaluation}
Bishal Lamichhane. 2023.
\newblock Evaluation of chatgpt for nlp-based mental health applications.
\newblock \emph{arXiv preprint arXiv:2303.15727}.

\bibitem[{Laurer et~al.(2024)Laurer, Van~Atteveldt, Casas, and Welbers}]{laurer2024less}
Moritz Laurer, Wouter Van~Atteveldt, Andreu Casas, and Kasper Welbers. 2024.
\newblock Less annotating, more classifying: Addressing the data scarcity issue of supervised machine learning with deep transfer learning and bert-nli.
\newblock \emph{Political Analysis}, 32(1):84--100.

\bibitem[{Lee et~al.(2022)Lee, Kang, Kim, and Han}]{lee2022detecting}
Daeun Lee, Migyeong Kang, Minji Kim, and Jinyoung Han. 2022.
\newblock Detecting suicidality with a contextual graph neural network.
\newblock In \emph{Proceedings of the eighth workshop on computational linguistics and clinical psychology}, pages 116--125.

\bibitem[{Lee et~al.(2023)Lee, Son, Jeon, Kim, and Han}]{lee2023towards}
Daeun Lee, Sejung Son, Hyolim Jeon, Seungbae Kim, and Jinyoung Han. 2023.
\newblock Towards suicide prevention from bipolar disorder with temporal symptom-aware multitask learning.
\newblock In \emph{Proceedings of the 29th ACM SIGKDD Conference on Knowledge Discovery and Data Mining}, pages 4357--4369.

\bibitem[{Li et~al.(2023)Li, Hammoud, Itani, Khizbullin, and Ghanem}]{li2023camel}
Guohao Li, Hasan Abed Al~Kader Hammoud, Hani Itani, Dmitrii Khizbullin, and Bernard Ghanem. 2023.
\newblock \href {https://openreview.net/forum?id=3IyL2XWDkG} {{CAMEL}: Communicative agents for ''mind'' exploration of large language model society}.
\newblock In \emph{Thirty-seventh Conference on Neural Information Processing Systems}.

\bibitem[{Liu et~al.(2023)Liu, Biester, and Mihalcea}]{liu2023improving}
Yujian Liu, Laura Biester, and Rada Mihalcea. 2023.
\newblock Improving mental health classifier generalization with pre-diagnosis data.
\newblock In \emph{Proceedings of the International AAAI Conference on Web and Social Media}, volume~17, pages 566--577.

\bibitem[{Luo et~al.(2023)Luo, Xie, and Ananiadou}]{luo2023chatgpt}
Zheheng Luo, Qianqian Xie, and Sophia Ananiadou. 2023.
\newblock Chatgpt as a factual inconsistency evaluator for abstractive text summarization.
\newblock \emph{arXiv preprint arXiv:2303.15621}.

\bibitem[{Malhotra and Jindal(2024)}]{malhotra2024xai}
Anshu Malhotra and Rajni Jindal. 2024.
\newblock Xai transformer based approach for interpreting depressed and suicidal user behavior on online social networks.
\newblock \emph{Cognitive Systems Research}, 84:101186.

\bibitem[{Schoene et~al.(2023)Schoene, Ortega, Amir, and Church}]{schoene2023example}
Annika~Marie Schoene, John Ortega, Silvio Amir, and Kenneth Church. 2023.
\newblock An example of (too much) hyper-parameter tuning in suicide ideation detection.
\newblock In \emph{Proceedings of the International AAAI Conference on Web and Social Media}, volume~17, pages 1158--1162.

\bibitem[{Shing et~al.(2018)Shing, Nair, Zirikly, Friedenberg, Daum{\'e}~III, and Resnik}]{shing2018expert}
Han-Chin Shing, Suraj Nair, Ayah Zirikly, Meir Friedenberg, Hal Daum{\'e}~III, and Philip Resnik. 2018.
\newblock Expert, crowdsourced, and machine assessment of suicide risk via online postings.
\newblock In \emph{Proceedings of the fifth workshop on computational linguistics and clinical psychology: from keyboard to clinic}, pages 25--36.

\bibitem[{Tang et~al.(2023{\natexlab{a}})Tang, Goyal, Fabbri, Laban, Xu, Yavuz, Kryscinski, Rousseau, and Durrett}]{tang-etal-2023-understanding}
Liyan Tang, Tanya Goyal, Alex Fabbri, Philippe Laban, Jiacheng Xu, Semih Yavuz, Wojciech Kryscinski, Justin Rousseau, and Greg Durrett. 2023{\natexlab{a}}.
\newblock \href {https://doi.org/10.18653/v1/2023.acl-long.650} {Understanding factual errors in summarization: Errors, summarizers, datasets, error detectors}.
\newblock In \emph{Proceedings of the 61st Annual Meeting of the Association for Computational Linguistics (Volume 1: Long Papers)}, pages 11626--11644, Toronto, Canada. Association for Computational Linguistics.

\bibitem[{Tang et~al.(2023{\natexlab{b}})Tang, Sun, Idnay, Nestor, Soroush, Elias, Xu, Ding, Durrett, Rousseau et~al.}]{tang2023evaluating}
Liyan Tang, Zhaoyi Sun, Betina Idnay, Jordan~G Nestor, Ali Soroush, Pierre~A Elias, Ziyang Xu, Ying Ding, Greg Durrett, Justin~F Rousseau, et~al. 2023{\natexlab{b}}.
\newblock Evaluating large language models on medical evidence summarization.
\newblock \emph{npj Digital Medicine}, 6(1):158.

\bibitem[{Williams et~al.(2017)Williams, Burnap, and Sloan}]{williams2017towards}
Matthew~L Williams, Pete Burnap, and Luke Sloan. 2017.
\newblock Towards an ethical framework for publishing twitter data in social research: Taking into account users’ views, online context and algorithmic estimation.
\newblock \emph{Sociology}, 51(6):1149--1168.

\bibitem[{Wu et~al.(2023)Wu, Irsoy, Lu, Dabravolski, Dredze, Gehrmann, Kambadur, Rosenberg, and Mann}]{wu2023bloomberggpt}
Shijie Wu, Ozan Irsoy, Steven Lu, Vadim Dabravolski, Mark Dredze, Sebastian Gehrmann, Prabhanjan Kambadur, David Rosenberg, and Gideon Mann. 2023.
\newblock Bloomberggpt: A large language model for finance.
\newblock \emph{arXiv preprint arXiv:2303.17564}.

\bibitem[{Xu et~al.(2023{\natexlab{a}})Xu, Yang, Lin, Wang, Zhou, Zhang, and Mao}]{xu2023expertprompting}
Benfeng Xu, An~Yang, Junyang Lin, Quan Wang, Chang Zhou, Yongdong Zhang, and Zhendong Mao. 2023{\natexlab{a}}.
\newblock Expertprompting: Instructing large language models to be distinguished experts.
\newblock \emph{arXiv preprint arXiv:2305.14688}.

\bibitem[{Xu et~al.(2023{\natexlab{b}})Xu, Yao, Dong, Gabriel, Yu, Ghassemi, Hendler, Dey, and Wang}]{xu2023leveraging}
Xuhai Xu, Bingshen Yao, Yuanzhe Dong, Saadia Gabriel, Hong Yu, Marzyeh Ghassemi, James Hendler, Anind~K Dey, and Dakuo Wang. 2023{\natexlab{b}}.
\newblock Mental-llm: Leveraging large language models for mental health prediction via online text data.
\newblock \emph{arXiv preprint arXiv:2307.14385}.

\bibitem[{Yang et~al.(2023{\natexlab{a}})Yang, Ji, Zhang, Xie, Kuang, and Ananiadou}]{yang2023towards}
Kailai Yang, Shaoxiong Ji, Tianlin Zhang, Qianqian Xie, Ziyan Kuang, and Sophia Ananiadou. 2023{\natexlab{a}}.
\newblock Towards interpretable mental health analysis with large language models.
\newblock In \emph{Proceedings of the 2023 Conference on Empirical Methods in Natural Language Processing}, pages 6056--6077.

\bibitem[{Yang et~al.(2023{\natexlab{b}})Yang, Zhang, Kuang, Xie, and Ananiadou}]{yang2023mentalllama}
Kailai Yang, Tianlin Zhang, Ziyan Kuang, Qianqian Xie, and Sophia Ananiadou. 2023{\natexlab{b}}.
\newblock Mentalllama: Interpretable mental health analysis on social media with large language models.
\newblock \emph{arXiv preprint arXiv:2309.13567}.

\bibitem[{Zhang et~al.(2023)Zhang, Liu, and Zhang}]{zhang-etal-2023-extractive-summarization}
Haopeng Zhang, Xiao Liu, and Jiawei Zhang. 2023.
\newblock \href {https://doi.org/10.18653/v1/2023.findings-emnlp.214} {Extractive summarization via {C}hat{GPT} for faithful summary generation}.
\newblock In \emph{Findings of the Association for Computational Linguistics: EMNLP 2023}, pages 3270--3278, Singapore. Association for Computational Linguistics.

\bibitem[{Zhang et~al.(2022)Zhang, Schoene, Ji, and Ananiadou}]{zhang2022natural}
Tianlin Zhang, Annika~M Schoene, Shaoxiong Ji, and Sophia Ananiadou. 2022.
\newblock Natural language processing applied to mental illness detection: a narrative review.
\newblock \emph{NPJ digital medicine}, 5(1):46.

\bibitem[{Zhang et~al.(2019)Zhang, Kishore, Wu, Weinberger, and Artzi}]{zhang2019bertscore}
Tianyi Zhang, Varsha Kishore, Felix Wu, Kilian~Q Weinberger, and Yoav Artzi. 2019.
\newblock Bertscore: Evaluating text generation with bert.
\newblock In \emph{International Conference on Learning Representations}.

\bibitem[{Zirikly et~al.(2019)Zirikly, Resnik, Uzuner, and Hollingshead}]{zirikly2019clpsych}
Ayah Zirikly, Philip Resnik, Ozlem Uzuner, and Kristy Hollingshead. 2019.
\newblock Clpsych 2019 shared task: Predicting the degree of suicide risk in reddit posts.
\newblock In \emph{Proceedings of the sixth workshop on computational linguistics and clinical psychology}, pages 24--33.

\end{thebibliography}
\bibliographystyle{acl_natbib}

\appendix
\clearpage

\begin{table*}\label{tab:highlight-prompt}

\section{Appendix} \label{sec:appendix}
\subsection{Prompt of Knowledge-aware Evidence Extraction} \label{sec:highlight-prompt}

\centering 
 \begin{tabular}{p{15cm}}
        \Xhline{1pt}
        \hline
        \parbox{15cm}{\centering \textbf{Knowledge-aware Evidence Extraction}}
          \\ \hline
        \hline
        \\
        \parbox{14cm}{You are a psychiatrist.

<Examples>

Suicide Risk Level: [\textbf{Example's Label}]

Post: [\textbf{Example's Post}]

Highlights: \\
1. [\textbf{Highlights of Example}]\\
...\\

Referring to the <Examples> , Identify the original phrases in the post that express or reference suicide risk factors and list them without modification about <Question>.
\\
\\
<Question>

Suicide Risk Level: [\textbf{Label}]

Post : [\textbf{Post}]

- The answer must be in numbering format [examples of formatting]\\
- Phrases should be included in the given <Question>'s post. \\
- You MUST refer <Question>'s given post.\\
- Highlight only necessary phrases, not full sentences.\\
- Select the parts that haves the suicide-related words like [\textbf{Suicide Word List}]\\
- Select as many phrases as possible related to suicide even a little.\\

Highlights:   }\\
\\
        \hline
        \Xhline{1pt}
    \end{tabular}
\end{table*}

\begin{table*}
\label{tab:prompt-summary}
\subsection{Evidence Summarization with LLM-based Consistency Evaluator} %
\subsubsection{Prompt of Evidence Summarization}
\label{sec:summary-prompt}
\centering
 \begin{tabular}{p{15cm}}
        \Xhline{1pt}
        \hline
        \parbox{15cm}{\centering \textbf{Extract-then-Generate}}\\
        \hline \hline
        \\
        \parbox{14cm}{
        You are a psychiatrist.\\
You are willing to do an abstractive summary about the evidence that shows the user is at {\textbf{[Label]}} suicide risk.\\
There are suicide risk assessment aspects when seeing Reddit posts.\\

[\textbf{GROUND TRUTH}]\\
The suicide risk level of this user is [\textbf{Label}]. \\
Here are Reddit posts and extractive evidence that supports the user is at {\textbf{[Label]}} suicide risk based on aspects. \\

[\textbf{Posts and Highlights}]
Regarding the user's posts and extracted evidence and aspects of suicide risk assessments, 
Explain why the user is at [\textbf{Label}] suicide risk.\\
The revised summary should include the information in the extractive evidence and aspects.\\
The summary should be shorter than 300 letters.\\
The summary MUST be less than 300 letters.\\

Summarized evidence explain: }\\
\\
        \hline
        \Xhline{1pt}
    \end{tabular}
\end{table*}

\begin{table*}
    \label{tab:prompt-summary}
    
\subsubsection{Prompt of Evaluating Consistency}
\centering
 \begin{tabular}{p{15cm}}
        \Xhline{1pt}
        \hline
        \parbox{15cm}{\centering \textbf{Consistency Evaluator}}\\ 
        \hline \hline
        \\
        \parbox{14cm}{
        You are an expertise psychiatry evaluator.\\
There are [\textbf{Label}] suicide risk user's posts and explain the reason for diagnosis based on posts.\\

[\textbf{Posts}]\\
Explain and summary of evidence: [\textbf{Summary}]\\
Score the following summary given the user posts concerning consistency from 1 to 10.\\
Note that consistency measures how much information the summary includes in the source posts. 10 points indicate that the summary contains only statements that are entailed by the source posts. 1 point indicates that the summary does not contain any word or statement that is entailed by the source posts.\\

Scores choices: from [1] to [10]\\

Give me a clear mark score and explain about it.\\
Keep the answer format\\
- Format: The score is [1] \\
to\\
- Format: The score is [10]\\

Scores: }\\
\\
        \hline
        \Xhline{1pt}
    \end{tabular}
\end{table*}

\end{document}